\pdfoutput=1

\documentclass[11pt]{article}

\usepackage{naacl2021}
\usepackage{booktabs}
\usepackage{times}
\usepackage{latexsym}
\usepackage{graphicx}
\usepackage{diagbox}

\usepackage[T1]{fontenc}

\usepackage[utf8]{inputenc}

\usepackage{microtype}

\title{Teaching NLP outside Linguistics and Computer Science classrooms: \\ Some challenges and some opportunities}

 \author{Sowmya Vajjala 
 \\ National Research Council, Canada \\
  \texttt{sowmya.vajjala@nrc-cnrc.gc.ca}
  }

\begin{document}
\maketitle
\begin{abstract}
NLP's sphere of influence went much beyond computer science research and the development of software applications in the past decade. We see people using NLP methods in a range of academic disciplines from Asian Studies to Clinical Oncology. We also notice the presence of NLP as a module in most of the data science curricula within and outside of regular university setups. These courses are taken by students from very diverse backgrounds. This paper takes a closer look at some issues related to teaching NLP to these diverse audiences based on my classroom experiences, and identifies some challenges the instructors face, particularly when there is no ecosystem of related courses for the students. In this process, it also identifies a few challenge areas for both NLP researchers and tool developers. 
\end{abstract}

\section{Introduction}
\label{sec:intro}

Until a few years ago, it was common to see NLP courses predominantly taught in either Computer Science or Linguistics departments, attended primarily by students from both the departments. In the past few years, there has been an increasing interest in NLP across disciplines. This is also reflected in the arrival of focused NLP courses such as "Clinical NLP"\footnote{\url{https://www.coursera.org/learn/clinical-natural-language-processing}} and "Introduction to Computational Literary Analysis"\footnote{\url{https://icla2020.jonreeve.com/}}.

This trend is by no means specific to NLP alone. There has been a growing interest in including courses related to computing/programming in many liberal arts and sciences departments over the past few years. \newcite{Guzdial-21} lists three reasons why many departments want such courses for their students: to support new discoveries through computational methods; to use new modes of interactive communication through apps, simulated environments etc.; to study the cultural, social and political influence of models and how to improve them. This need resulted in the creation of data science programs open for all students from various colleges/departments. Naturally, the introduction of such programs started discussion around what should be included in such \textit{generic} data science introductory curricula  \cite{Krishnamurthi.Fisler-20}. An introductory course in NLP is commonly offered either as an elective or as a part of the main coursework in most such data science course/minor/certificate programs in universities. Such a course is also a common component in industry facing certification programs offered outside of university settings. 

However, popular textbooks and course materials on NLP are not created taking these diverse audiences and their motivations into consideration. They cover a range of topics in depth, requiring a deeper technical background that students coming from diverse academic backgrounds may not possess. While there are some recent efforts to write books focused to specific groups of students  \cite{Jockers-14,Hovy-20}, they tend to focus on a smaller subset of topics within NLP. 

Further, most courses and textbooks focus on the algorithms, with much less attention given to other essentials such as data collection, text extraction, pre-processing and other practical issues one will encounter when working on new research problems (and datasets), or while deploying NLP systems in some application scenario. Considering that many students take these courses with goals unrelated to performing NLP research later (e.g., using NLP in their disciplinary research, working for a software company, etc.), this lack of appropriate materials can be seen as a potential gap between current NLP teaching and what is required. Additionally, while research progress feeds into the development of course syllabi, we don't see the opposite i.e., teaching experiences informing NLP research. 

In this background, this paper summarizes my experiences with teaching multiple NLP courses to a diverse student body (STEM, humanities, social sciences) both at undergrad and graduate level, and identifies some challenge areas for classroom instruction. At the same time, it also identifies some relatively understudied problems in NLP research and some issues to consider for tool developers. In short, the contributions of the paper can be summarized as follows:
\begin{itemize}
    \item It identifies the challenges of teaching a range of student audiences outside of linguistics and computer science departments. 
    \item It identifies a few challenge areas for NLP research and practice, which should be addressed to further the use of NLP methods and techniques beyond computer science and related disciplines. 
\end{itemize}

 The rest of this paper is organized as follows: Section ~\ref{sec:relw} gives a brief overview of existing work on teaching NLP to put this paper's contributions in context. Section ~\ref{sec:ourexp} describes the NLP courses I taught with diverse goals and for diverse audiences, in detail. The next section then identifies some challenges in terms of teaching for all these different contexts (Section ~\ref{sec:gap}). Section ~\ref{sec:challenge} elaborates on how these teaching experiences help us identify some challenge areas for NLP  researchers as well as tool developers. Section ~\ref{sec:concl} summarizes and concludes the paper. 

 \section{Teaching NLP - A short review}
\label{sec:relw}
Since the first Teaching NLP workshop almost two decades ago, there has been some discussion in the community on various issues related to NLP pedagogy through the TeachCL/TeachingNLP workshop series,\footnote{\url{https://www.aclweb.org/anthology/venues/teachingnlp/}} and other events such as Teach4DH.\footnote{\url{https://teach4dh.github.io/}}. Published work on teaching NLP can be broadly classified into four groups:
\begin{enumerate}
    \item Sharing insights about building new CL programs in various intra- and inter-disciplinary contexts \cite[e.g.,][]{Lonsdale-02,Dale.Diego.ea-02,Koit.Roosmaa.ea-02,Baldridge.Erk-08,Zinsmeister-08,Reiter.Schulz.ea-17}
    \item Focused discussion about the challenges in the design of single courses, sometimes targeting a broader/diverse audience \cite[e.g.,][]{Liddy.McCracken-05,Xia-08,Madnani.Dorr-08,Agarwal-13,Navarro-17} 
    \item Development and usage of specific tools/games/strategies for teaching NLP \cite[e.g.,][]{Bontcheva.Cunningham.ea-02,Lin-08,Bird.Loper.ea-08,Hockey.Christian-08,Levow-08,Barteld.Flick-17}
    \item Discussion around the design of linguistic problems for North American Computational Linguistics Open Competition  (NACLO)\footnote{\url{https://nacloweb.org/}} and other events \cite[e.g.,][]{Bozhanov.Derzhanski-13,Littell.Levin.ea-13}. 
\end{enumerate}

However, to my knowledge, there hasn't been much discussion on the gap between what we learn in the classroom versus how you use it later (whether in research or practice), how we should adapt the syllabus depending on the audience for the course, and on how teaching experiences can potentially inform NLP research and practice. I addressed these issues in this paper, based on my experiences with teaching NLP. 

\section{Description of NLP Courses}
\label{sec:ourexp}
This section describes the teaching scenarios in which I taught NLP between 2016-2021, which form the basis for the observations discussed in this paper. Full semester courses described below were taught at an American university during 2016-18, and the online, compact courses were taught at two German universities during 2020-21. All classrooms were small in size ($<$ 30 students) and there were no teaching assistants. Inspired by \newcite{Bird-08}'s classification of student's background and goals for using NLTK \footnote{Table 3.1 in \url{http://www.nltk.org/book/ch00.html}}, and \newcite{Fosler-Lussier-08}'s grouping of students taking NLP courses, a grouping of my NLP teaching contexts, is shown in Table ~\ref{tab:1}, taking student background and course goals into consideration. The "General" courses, as the name indicates, target a broader audience, while "Focused" courses were created to address specific student needs. 

\begin{table*}[t]
  \centering
  \begin{tabular}{|p{30mm}|p{35mm}|p{35mm}|}
  \hline
     & \textbf{General} & \textbf{Focused} \\ \hline
    \textbf{Undergrad} & A non-technical course intended to give an overview of language and computers, open for all disciplines & An introductory course taught as a part of data science curriculum for Liberal Arts and Sciences (LAS) majors\\ \hline
    \textbf{Grad/Advanced \hspace{2mm} undergrad} & Two general NLP courses with a focus on various algorithms and applications & Two introductory NLP courses for specific graduate groups (applied linguists and economists)\\ \hline
  \end{tabular}
  \caption{NLP Teaching Contexts in terms of student groups}
  \label{tab:1}
\end{table*}

 The rest of this section presents a detailed overview about the courses taught under these groups. I hope to achieve the following goals through this long overview:
 \begin{enumerate}
     \item share some insights about what to include/exclude in the syllabus/exercises, when we are developing a new NLP course for a non-traditional audience
     \item provide a useful context to the challenges that will be discussed later in the paper. 
 \end{enumerate}

\subsection{Undergrad-General (U-Gen): }
I taught one course\footnote{\url{https://github.com/nishkalavallabhi/LING120-Fall2017/}} which could be classified as a general undergraduate course that is open for all. This is based on the textbook by \newcite{Dickinson.Brew.ea-12}, which is used to teach several such "Language and Computers" courses around the world. Students in this course came from all years of undergraduate curriculum, but were dominated by freshmen and sophomores from Sciences and Engineering. The course had an opt-in programming component for enthusiastic students, but otherwise, generic enough that all freshmen undergraduate students from any discipline can follow. 

\paragraph{Syllabus: } The topics covered followed the book's structure, with more contemporary information. For example, the topic "tutoring systems" included a discussion of software such as DuoLingo, and the topic "dialog systems" included discussion on virtual assistants such as Siri, Cortana etc. The focus in using these tools in this classroom is to explain how such systems work, and evaluate their performance in real world contexts, rather than on teaching how to build such systems. Two sessions were spent on giving a non-technical overview of NLP with some discussion on recent trends and on the broader impact of NLP on other areas of study. 

\paragraph{Evaluation: } The assignments in this course required students to explore a few existing NLP tools/demos (e.g., corenlp.run) or browse existing corpora (e.g., corpus.byu.edu). The students also did a group presentation that involved using a readily available day to day software which has some NLP component and summarizing its performance with concrete examples and qualitative measures. The final exam for this course had two parts: the first part required the students to write a report performing a error analysis of two commercial systems doing the same task (e.g., machine translation, speech recognition, etc.) and the second part asked them to write a perspective essay on the impact of language technologies on the society.

\subsection{Undergrad-Focused (U-Focused)} 
I taught one course\footnote{\url{https://github.com/nishkalavallabhi/LING410X-Spring18}} which was one of the electives in an undergraduate data science minor program. This was open to students from all departments under the school of Liberal Arts \& Sciences. It was taught twice and attracted students from a wide range of disciplines including, but not limited to: Literature, Sociology, Journalism, Management, World Languages, Linguistics and Computer Science.  

\paragraph{Syllabus: } The goal of this course was to introduce students to methods of discovering language patterns in text documents and applying them to solve practical text analysis problems in their disciplines. The course required students to write a few programs, although the knowledge of programming was not a pre-requisite. Since many of these students were getting familiarized with using R as a part of their curricula (for statistics courses), and I viewed it as a language in which they can make some progress without having to gain expertise in programming simultaneously, the course was taught in R. Relevant programming concepts and data structures were introduced in the context of the topics of the course, on the fly. \newcite{Jockers-14} was used as the main textbook for this course, since it assumed no programming background from its target audience (Literature students), and it used R. 

In terms of course organization, a few sessions were spent on introducing students to the basics of installing and using R, specifically with the goal of working with textual data in mind. The next topic discussed methods to collect, extract and clean texts/corpora. This was followed by teaching about doing basic exploratory corpus analysis along with keyword and key phrase extraction approaches. The next topics focused on text classification and topic modeling - both of which are the most commonly used methods with textual data across disciplines. The final topic for the course discussed various means of visualizing textual data. 

\paragraph{Evaluation: } The course included assignments that followed the topic structure, and required students to write small R programs to extract text patterns, scrape data from different forms of documents (e.g., webpages, twitter), extracting keywords, ngrams etc., following step by step process making alterations to pre-written code for training a text classifier and a topic model, and building basic visualizations of textual data (e.g., word clouds, dispersion plots etc). All assignments relied on learning to use existing R libraries instead of focusing on building everything from scratch. The students did a group presentation which involved visualizing textual data. The final exam consisted of doing a small project and submitting the term paper. The students were required to pick an interesting dataset for a problem they encountered in their own discipline, and use one of the methods they learnt in the course. 

\subsection{Grad/Advanced undergrad-General (G-Gen)}
\label{subsec:advgeneric}
I taught two courses that are the closest to a typical Natural Language Processing course taught in universities across the world: the first course followed the standard structure in traditional textbooks, and the second course focused specifically on how to do NLP in the absence of large annotated datasets. Students in both courses primarily consisted of advanced undergrad or graduate students coming from linguistics and computer science, with varying degrees of background in programming, linguistics, and computer science. All students passed at least one programming course prior to attending these courses. 

\paragraph{Statistical NLP:} The course's objective was to teach some of the common algorithms and techniques that form the foundation for modern day NLP. Accordingly, starting with regular expressions and language models, the topics we dealt with included part of speech tagging, parsing, discourse, information extraction, text classification and other NLP applications, and ended with introducing text embedding representations along with an overview of neural network architectures. \newcite{Jurafsky.Martin-08} and \newcite{Goldberg-17} were used as the prescribed textbooks. Assignments focused on implementing some of the popular NLP algorithms from scratch, and final project involved modeling one of the common tasks such as text classification or information extraction using existing datasets and producing a technical report describing the same. 

\paragraph{NLP without data\footnote{\url{https://github.com/nishkalavallabhi/SfSCourseJan2021}}}: The objective of this course was to address a real world scenario that is not typically addressed in NLP courses - how do we apply NLP methods in the absence of annotated training data. Hence, after giving a broad overview of NLP and its uses both in real world and for other disciplines, and discussing in detail about NLP system development pipeline and text representation, the course focused on the following topics:
\begin{enumerate}
    \item Corpus collection, text extraction and exploratory analysis
    \item Automatically labeling data and performing data augmentation
    \item Working with small datasets and transfer learning
\end{enumerate}

This course was taught remotely, as a compact, intensive 1 month online course in January 2021 ($~$ 9 hours per week, for 3 weeks) due to the current pandemic situation. It primarily relied on a collection of blog posts, research papers, code tutorials and python notebooks from various sources, as no available textbook specifically addressed this topic. It had two assignments, which required students to explore the usage of existing NLP tools to perform given tasks and to "generate" labeled data for information extraction, and write up a report evaluating their approaches with some error analysis. There was a group presentation, where students had to pick from a selected collection of recent research articles on the course's topics. Finally, there was an optional term paper (for extra credits), where the students can pick a resource scarce NLP scenario and apply what they learnt in this course to address it.  

\subsection{Grad/Adv.undergrad-Focused (G-Focused)}
I taught two introductory NLP courses that were specific to graduate students - one for applied linguistics Masters/PhD students and the other for Economics Masters/PhD students. 

\paragraph{NLP for Applied Linguists}
Applied linguistics graduate program in the university where this course was taught consisted of students studying corpus linguistics, computer assisted language teaching/learning, and technical communication. Since the use of language processing tools in these areas is increasing day by day, the goal of this course was to teach some text processing methods that can be directly applicable in their dissertation research. Accordingly, the course introduced various NLP techniques from regular expressions to using parsers, in the context of technologies for language learning and corpus analysis e.g., spelling/grammar checkers, pattern extractors for corpus analysis etc. \newcite{Bird.Loper.ea-08} and \newcite{Jurafsky.Martin-08} were used as the textbooks for this course, along with \newcite{Church-94}.

The course had five assignments which involved writing small programs covering the topics of the course, and a group project, followed by a one-to-one oral exam to assess student understanding of the relevance of the course to their area of study. 

\paragraph{NLP for Economists\footnote{\url{https://econnlpcourse.github.io/}}: } This was originally planned to be a one week intensive course, but was taught online over three weeks due to the pandemic situation in Fall 2020. It also included instruction on Python fundamentals. The syllabus for the topic comprised of four broad topics:
\begin{enumerate}
    \item Introduction: An overview of NLP and its usefulness in Economics; Python fundamentals
    \item Python \& textual data, which focused on data collection, text extraction, pre-processing and text representation
    \item NLP methods for economics, covering exploratory corpus analysis, text classification, topic modeling, and giving an overview of others such as information extraction and text summarization
    \item NLP in Economics - research paper readings and discussion
\end{enumerate}

The students had to do 3 assignments covering the first three topics which involved writing small python programs to use existing tools and write an analysis of how they work for their domain data. They also did a group presentation picking a paper that used NLP methods to address research questions in economics, chosen most often from their own disciplinary journals, instead of NLP conferences. Students also had to submit a term paper  which involved creation of a problem statement describing a new economics problem that can be addressed using NLP methods. No specific textbook was used for the course, although several recommendations were listed in the syllabus. The course videos and slides, and links to publicly available online content were the primary materials for the course, as there was no appropriate textbook available to suit the needs of this audience.  

As it can be seen from all the above course descriptions, the courses addressed a range of audiences, and accordingly, differed in the way the course was organized as well as the topics that were covered. Most of these courses can be called "non-traditional" NLP courses, considering their contents and intended audience. In the following section, I will elaborate on some of the general and course specific challenges I faced in the design and delivery of these courses. 

\section{Challenges for Teaching}
\label{sec:gap}
While all the courses received generally good student feedback towards the end, there are several issues that could have been managed better. Some of these issues arise because I am teaching a diverse, non-conventional NLP audience, and hence, we don't have readily available solutions yet. Each course, of course, comes with its own challenges. In this section, I will discuss some of the more general teaching challenges I faced across all courses, and how I addressed them. 

\paragraph{Student goals and Course Contents: } For many of the courses described in the previous section, the students did not have an ecosystem of related courses in their curriculum because there are no NLP focused research groups or teaching programs in the university. For all except \textit{G-Gen} courses, any NLP course is potentially a one off course that covers programming, NLP, and anything remotely concerned with computing for the students. The student goals accordingly are related to either just fulfilling a course requirement, or gaining some quick actionable insights that they can use right away, or show relevant skills when they apply for a job soon. In this background, I found it particularly challenging to adapt the syllabi such that they get readily usable practical skills, along with a solid foundation to explore the topics further on their own if needed. 

An approach that seemed to work is to tie each concept of the course to a practical use case (either a software application or some research problem in another discipline) and give assignments that are closer to their real-world scenarios. In the \textit{U-Gen} course, group activities involving apps the students regularly use such as Duolingo, Google Search, Siri etc generated a lot of interesting questions in the class. In the \textit{G-Focused} courses, providing an overview about relevant disciplinary research that uses NLP, and encouraging discussions about how the students can use NLP in their research turned out to be useful. The NACLO exercises helped to introduce the challenges while working with textual data, in a way that holds students' attention and generated interesting discussions, in all the four cells in Table~\ref{tab:1}. 

\paragraph{Faculty goals and Course Contents: } As mentioned above, many students lacked the eco-system of related courses (all except \textit{G-Gen}). Thus, a lot of surrounding topics that are not typically a part of NLP courses had to be covered in these classes. Specifically, these topics revolved around concepts of software programming, and the mathematics needed to understand some of the basic NLP methods. Two questions I constantly grappled with in terms of my own teaching goals for \textit{U-Focused} and \textit{G-Focused} courses courses were - how much mathematics/programming/linguistics should be included? how can I encourage good programming/software engineering practices while still focusing on teaching the students to solve NLP related problems? 

For the first question, keeping mathematics and linguistics to the bare minimum, situating all programming related exercises in the context of NLP/textual data, and providing additional resources/tutorials for those interested in knowing more math/linguistics behind NLP helped . For the second question, I tried to write clean, well-commented code for classroom examples as much as possible, and showed variations of writing the same piece of code, on top of the exercises in the prescribed textbook, discussing why we may chose one over the other. I also encouraged students to review each other's submissions and post discussions about programming in the forum for some of these courses. Both these teaching strategies created some awareness about programming practices among the students. 

However, there were always students who wanted more/less of math/linguistics/programming during the classroom session itself, or in the form of assignments, especially in \textit{G-Gen} and \textit{G-Focused} courses. Some students felt enough challenged, some were overwhelmed. This issue is by no means specific to my experiences, and has been documented in past work on teaching NLP too \cite{Brew.Dickinson.ea-05,Koit.Roosmaa.ea-02}. I addressed this by providing optional, additional (ungraded) programming exercises and reading materials in all the courses. 

In terms of programming practices, I found it particularly challenging to introduce the idea of version control for code for students in \textit{U-Focused} and \textit{G-Focused} classrooms. The students without prior background in any form of programming found it difficult to understand Git. I emphasized its importance and spent a small amount of time discussing why version control is useful, and how to do it, and provided a few tutorial references. While none of the students used version control during the courses, I hope that as they gain more experience, they will understand its relevance and adopt in practice. 

\paragraph{Textbooks and the real world: } If we ask ourselves - what will the students do with what they learn in these courses, provided they manage to stick to NLP?, we can think about three options: a) pursue further studies/research focused on NLP b) pursue further studies/research in their own discipline, using NLP methods in their work c) work in a company on NLP projects. Among these, we can safely assume that the last two are the most likely scenarios for the audience I taught. 

In my experience as a software engineer and as a data scientist in industry, and while collaborating with researchers from other disciplines, some of the most common issues I encountered are:
\begin{itemize}
    \item How do we collect and label the data needed to train NLP models?
    \item How can we do a clean and accurate extraction of text from various file formats?
    \item What are the efficient ways of selecting models going beyond standard intrinsic evaluation measures (e.g., considering external measures, deployment costs, maintenance issues etc)
\end{itemize}
Most of this issues are also commonly experienced by NLP researchers, if they work on a new problem or on creating a new corpus resource. Yet, none of these issues are discussed even cursorily in available/standard NLP textbooks, which means that the students have to rely on a lot of online blogs and other resources. 

I addressed these issues by including topics such as: what does a NLP system development pipeline look like? and how is NLP used beyond CS research and software development scenarios? in the introductory "NLP overview" sessions itself in \textit{G-Gen} classrooms, introducing the students to the challenges one faces at each step. In the rest of the course too, I addressed these issues in more detail as needed, providing code examples, and including real-world case studies. Particularly, one course described in this paper, "NLP without annotated dataset", entirely focused on the first issue mentioned earlier. 

In \textit{U-Focused} and \textit{G-Focused} classrooms, I added a detailed overview of how NLP is used in various disciplines as a research method, and organized student group discussions with contemporary research papers in these disciplines. Even in the relatively simpler \textit{U-Gen} course scenario, we ran into the issue of the existing textbook being outdated, which required me to look for new materials on the topic. As mentioned earlier, I addressed this issue by introducing discussions and assignments on more contemporary developments about the textbook's topics. 

I found it particularly challenging to cover all of these aspects in one course, while also giving enough background in core NLP topics, though. In future, it would be useful to develop some structured reading material/tutorials/workbooks that can serve as goto sources on these topics, rather than asking the students to just explore on their own from a wide range of available material on the web.

\paragraph{Graded Resources: } 
Another recurring challenge I ran into relates to the availability of appropriate textbooks. While I found introductory textbooks that suited some of the classes, the students in the \textit{G-Focused} group repeatedly mentioned the lack of a progressively difficult self-learning path. As they rightly pointed out, we either have introductory books which gave a basic idea of selected topics \cite[e.g.,][]{Jockers-14}, or very advanced textbooks \cite[e.g.,][]{Jurafsky.Martin-08}, but nothing in between. The issue of lack of resources was also highlighted by \newcite{Hearst-05} in the context of teaching an applied NLP course. Although this was to some extent addressed by some of the more practically oriented books published by O'Reilly Media\footnote{\url{https://www.oreilly.com/}} and Manning Publications\footnote{\url{https://www.manning.com/}}, they still addressed industry facing scenarios, which did not always account for these students' needs. 

\paragraph{Teaching diverse audience: } All except the \textit{G-Gen} courses described in this paper had either students coming from diverse disciplines, or a homogeneous group belonging to a non-STEM discipline. In such cases, it is important that the students understand the relevance of the course to their own discipline. This can be challenging, if we, as the instructors, don't know what are some interesting challenges in the various disciplines. The need to target courses to the student background was also discussed in the past in \newcite{Baldridge.Erk-08}. To this end, I had several informational chats with the faculty members from these disciplines, to understand the use cases for NLP in their research, apart from reading research that used NLP methods in the respective disciplines. This was definitely useful in making NLP more relevant to the student groups. Co-teaching such courses with faculty from other disciplines is an idea to explore in future, provided the class is homogeneous, and both instructors are willing to invest time and energy into learning the methods of the other discipline. 

These are some of the challenges I faced in teaching NLP and how I addressed them. However, all these issues are by no means completely answered, and more discussion is needed in this direction. Teaching NLP workshops, and sharing of teaching resources (and anecdotes) can be a starting point towards addressing these issues. 


 \section{Challenges beyond the Teaching context}
 \label{sec:challenge} 
 Generally, we see some discussion about how research informs teaching and the choice of topics in a course. We may notice the evolution of the standard Natural Language Processing course over the past two decades, in terms of the topics covered\footnote{NLP Pedagogy interviews by David Jurgens and Lucy Li\url{shorturl.at/bntR5}}. We also see a lot of discussion around challenges encountered in teaching itself, as seen in the Teaching NLP papers in the past. However, when teaching to \textit{outside} audience, we can uncover hitherto under-studied research questions that are potentially more relevant while doing NLP outside of computer science and linguistics. I will focus on such issues in this section, by splitting it into two groups: NLP research and tool development.

 \subsection{NLP Research}
 The questions raised by the students in \textit{G-Focused} courses identified some under-studied issues in contemporary NLP research, which are described below:
 
 \paragraph{Predictions and Causality: } In NLP, we generally work with various representations of textual data, and build predictive models. Though there is an increasing body of research on understanding the predictions, it is not common to see a causal analysis. However, in some disciplines, such causal relationship is important for any prediction. While teaching an economics classroom, we repeatedly ran into these discussions about drawing causal inference for a text classifier or a topic modeling decision. These led a student to a question - will NLP ever be really useful in economics research beyond being a fancy new technique? 
 
There is some existing work that uses causal analysis along with NLP methods in economics literature \cite{Gentzkow.Kelly.ea-19}, but there is not much within NLP research in that direction, focusing on specific applications. Although there is some recent interest among NLP researchers on causal inference \cite{Veitch.Sridhar.ea-20,Keith.Jensen.ea-20}, we do not yet have off the shelf teaching resources to incorporate such aspects into the classroom, to my knowledge. More NLP research and inter-disciplinary collaborations in this direction may lead to the creation of the much needed teaching and software resources to address this important issue in future. 
 \paragraph{Text as secondary data: }  Unlike typical NLP focused problems, text is a form of secondary data for many other disciplines (e.g., economics, again). Thus, their expectation is that the primary source of information for model predictions should come from the primary data sources. While discussing feature representations for text, we repeatedly ran into the issue of how to effectively combine these different feature representations coming from primary and secondary sources. It is certainly possible to concatenate representations or create multimodal representations and let the model figure out feature importance, as we commonly do in NLP research. However, the students questioned this approach and asked for methods to develop models such that their primary feature representations were given importance over textual representation. 
 
 I am not aware of a commonly used approach for the same,  that can be incorporated directly in a course through theory or practical exercises. However, I believe we should acknowledge that the students are more familiar with research methods from their own disciplines and want to use NLP within that framework, rather than completely switch to "the NLP way" of building models. New research on using text representation as a secondary feature vector along with primary features may be a step in the direction of addressing this issue. 

 \paragraph{Construct validity of text representations: } Construct validity refers to whether a feature actually measures what it is supposed to measure. In language testing literature, construct validity is frequently studied in the context of automated language assessment software. In the Applied Linguistics graduate course, we discussed  automatic essay scoring and spelling/grammar correction systems briefly, as use cases of NLP in language assessment and teaching. During these discussions, the students, who typically came from a language assessment background, questioned the lack of traditional steps such as exploratory analyses to understand the corpus, and evaluating the construct validity of the feature representations we use in NLP. While we see some discussion around these issues in the context of real world NLP system development \cite[e.g.,][]{Sheehan-17,Beigman-Klebanov.Madnani-20}, we don't see much discussion about such issues in research describing the development of new NLP methods or in NLP textbooks. As we see NLP being used more and more outside its typical contexts, perhaps, it is time for us as NLP researchers to consider these issues while analyzing models and their performance. 

All these are important problems  beyond teaching NLP, for NLP researchers in general, and in the inter-disciplinary research context in particular. As \newcite{Connolly-20} suggests, supplementing computing related courses with methods, and perspectives from social sciences may give new insights for NLP researchers into addressing these issues. This will benefit teaching NLP contexts beyond the traditional audience, and also enrich NLP research in future.

 \subsection{Tool Development}
 In all the courses taught outside of classrooms dominated by STEM students (i.e., all except \textit{G-Gen}), I frequently ran into the issue of insufficient documentation for NLP tools the students want to use. From my personal experience, this situation is constantly improving. Yet, it is far from ideal. Despite being actively involved with using various NLP tools in daily life, it is not uncommon for many of us to face some challenges with software usage.
 
 This is even harder for those without that background with DIY software. Especially, tools that work across all commonly used operating systems are not very easy to find, but are essential when we want to use them in classrooms. It was particularly challenging even as an instructor, while teaching the data science minor course which used R. Since the students preferred to use their own machines, it was not possible to enforce a common operating system/library versions setup. So, I addressed this issue by guiding the students with links to videos on installation of various tools on all common operating systems where possible, and using alternative libraries where this did not work. Other issues the students raised were about the lack of proper graphical user interfaces and visualization methods for NLP tools. While I could not offer any clear solutions for these issues, tool developers should perhaps keep a broader, potentially non-technical users in mind when they release and document their tools in future. 
 
 What is described in this section is merely a snapshot of some of the teaching NLP challenges that go beyond the teaching context, and need the attention of other members of the  NLP community - researchers and tool developers. The list mentioned in this section is by no means exhaustive, and more discussion is needed in this direction especially in the current situation where NLP is taught and used by many disciplines beyond linguistics and computer science. 
 
 \section{Conclusion}
 \label{sec:concl}
 In this paper, I summarized my experiences with teaching NLP courses for diverse groups of undergraduate and graduate students and identified a few challenge areas for teaching such courses. I also discussed few challenge areas for NLP researchers and tool developers, addressing which can help improve both teaching NLP and the ease of applying NLP in research areas beyond linguistics and computer science. It should be acknowledged though, that this discussion is more qualitative than quantitative in nature. Perhaps doing a survey of the students of such courses a couple of years later about how they are using NLP compared to what they learnt in a classroom could be one way to measure these observations in a more quantitative manner. I hope that this paper will contribute to the growing body of work on NLP Teaching, and also lead to  further discussion on an increased focus on the inter-disciplinarily relevant aspects of NLP teaching, research and practice. 
 
 \section*{Acknowledgements}
Firstly, I thank the workshop committee for organizing this workshop. Comments from all the three anonymous reviewers, and my colleagues - Rebecca Knowles, Gabriel Bernier-Colborne and Taraka Rama were immensely useful in bringing the paper from the first draft to the final version - I thank them all for their time and thoughts on this paper. Finally, I thank the students in all these courses, and the three universities (Iowa State University, USA; Ludwig Maximilian University of Munich, Germany; Eberhard Karls University of T\"ubingen, Germany) that gave me the opportunities to teach them. 
 
 
\bibliography{anthology,custom}
\bibliographystyle{acl_natbib}




\end{document}